\documentclass[11pt]{article}

\usepackage[preprint]{acl}

\usepackage{times}
\usepackage{latexsym}
\usepackage{amsmath}
\usepackage{amssymb}
\usepackage{makecell}
\usepackage[T1]{fontenc}

\usepackage[utf8]{inputenc}
\usepackage{subcaption}
\usepackage{microtype}
\usepackage{booktabs}
\usepackage{multirow}
\usepackage{tabularx}
\usepackage{array}
\usepackage{inconsolata}

\usepackage{graphicx}
\newcolumntype{Y}{>{\raggedright\arraybackslash}X}

\usepackage{textcomp}
\newcommand{\ndminus}{\mathbin{\text{\textminus}}}

\title{When to Plan, When to Polish: Noise Level as a Granularity Axis for Diffusion Language Models}

\author{Peihong Li\textsuperscript{*} \\
  Washington State University \\
  \texttt{peihong.li@wsu.edu} \\\And
  Yuanjie Shi\textsuperscript{*} \\
  Washington State University \\
  \texttt{yuanjie.shi@wsu.edu} \\\And
  Yan Yan \\
  Washington State University \\
  \texttt{yan.yan1@wsu.edu}
  }

\begin{document}
\maketitle

\begingroup
\renewcommand{\thefootnote}{\fnsymbol{footnote}}
\footnotetext[1]{Equal contribution.}
\endgroup

\begin{abstract}
Standard tokenwise diffusion LMs keep training corruption and inference commitment at token granularity throughout denoising. At high noise, this leaves scattered local fragments rather than coherent evidence, making it hard to form early coarse structure, exactly what planning-sensitive generation requires.
Hierarchical planning methods add coarse stages to separate planning from wording, but they need extra planners, block latents, or two stage designs. We propose Noise Dependent Granularity Control (NDGC), a single-level diffusion method that uses the noise level as a granularity cue. NDGC aligns training exposure and inference commitment with denoising progress. High noise steps use coherent token groups to support early meaning commitment, while low noise steps return to token level refinement. This creates planning like coarse to fine denoising without an explicit planner or hierarchical architecture. Across controlled tests, ablations, and WritingPrompts, NDGC shows earlier skeleton formation, better ordered recovery, and healthier outputs.
\end{abstract}

\section{Introduction}

Discrete diffusion language models generate text by denoising a masked sequence as noise decreases \cite{sahoo2024simple,shi2024simplified,lou2023discrete,austin2021structured,li2022diffusion}.
Starting from a corrupted target region, the model predicts clean tokens and commits parts of the sequence step by step.
This form is useful for conditional long form generation because the model can revise the whole target, rather than only extend a prefix from left to right \cite{lee2018deterministic,ghazvininejad2019mask,wang2019bert,stern2019insertion,gu2019levenshtein}.
As denoising reduces uncertainty over time, the noise level can serve as a granularity cue: high-noise steps favor coarse semantic structure, and low-noise steps favor token-level wording refinement \citep{he2023diffusionbert,zheng2023reparameterized}.



\begin{figure}[t]
\centering
\includegraphics[width=0.95\linewidth]{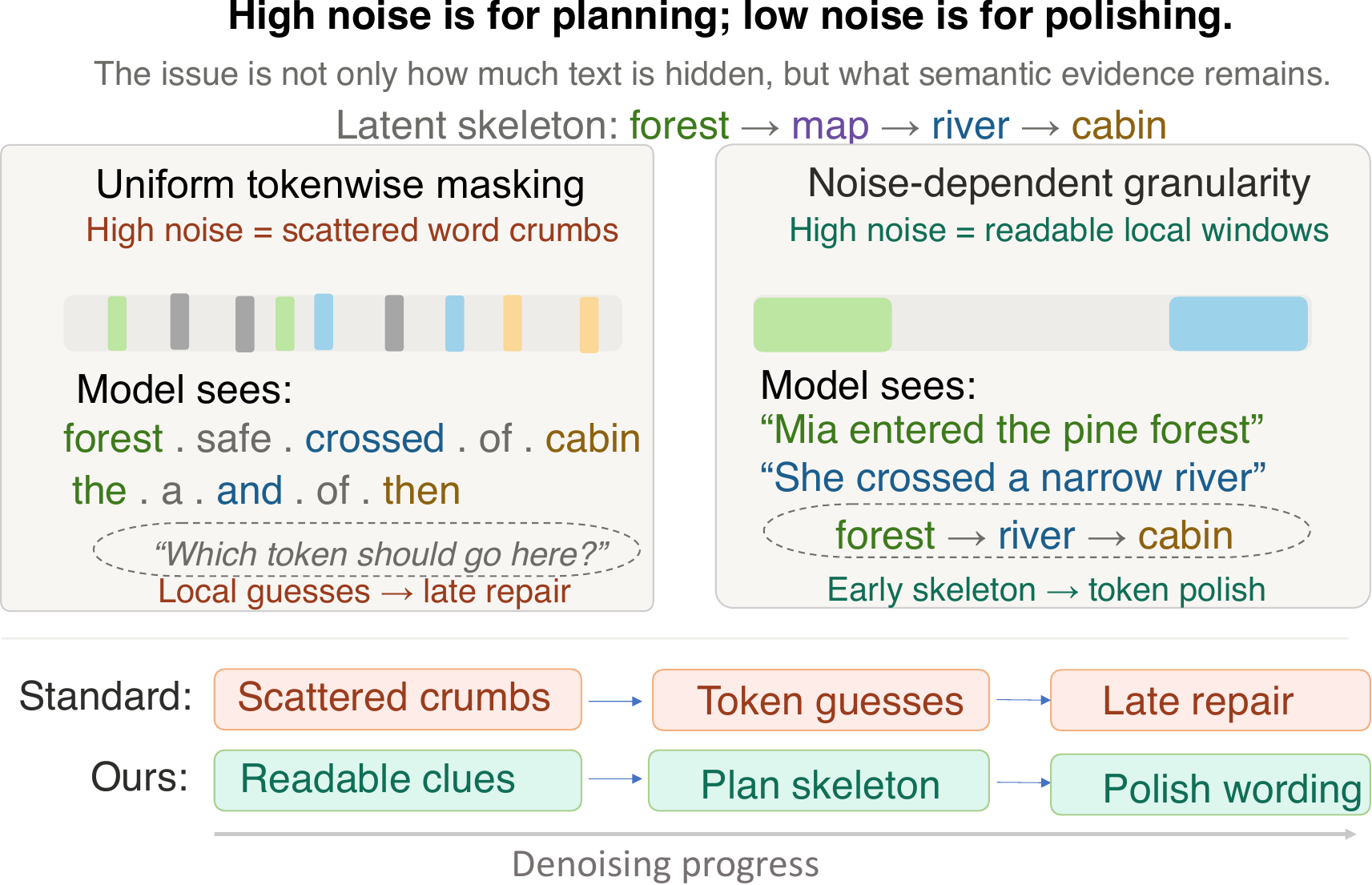}
\caption{
\textbf{Intuition behind noise-dependent granularity.}
Uniform tokenwise masking keeps the same token-level granularity at high noise,
leaving scattered word clues and making early denoising rely on local guesses.
NDGC keeps the expected corruption level fixed but changes the granularity:
high-noise states expose coherent token groups and commit groups during sampling,
so early denoising can form a coarse skeleton before later token-level polishing.
}
\label{fig:fig1}
\end{figure}

However, a high-to-low noise trajectory does not automatically create
coarse-to-fine generation. In standard tokenwise diffusion LMs, both
training corruption and inference commitment are still done token by token.
Thus, high noise controls how many tokens are hidden, but not whether the
model works at a topic, span, or token scale.\cite{zhang2025flexible,gwak2025reward,koh2025conditional,zhou2024diffusion}.
At high noise, independent token corruption leaves isolated fragments, so the model predicts tokens from weak local clues rather than coherent passages \cite{gong2023diffuseq}.
Figure~\ref{fig:fig1} illustrates this failure mode: high-noise tokenwise masking leaves scattered evidence, turning early denoising into local guessing rather than structure formation.
This is a poor fit for generation that needs planning.
Early steps should set topics, discourse order, or scene structure, while later steps should refine wording \cite{fan2018hierarchical,yao2019plan,fan2019strategies,yang2022re3,goldfarb2020content}.

Hierarchical planning methods add coarse stages to separate planning from wording, but they need extra planners, block level latents, or two stage designs \cite{fan2018hierarchical}.
Plan and Write systems build an outline before writing the full text, and block level diffusion methods separate coarse structure from token detail through extra stages \cite{yao2019plan,arriola2025block}.
As a result, they do not answer whether the original high-to-low noise trajectory of a single-level diffusion LM can itself support early planning and late token refinement.
We ask a smaller question: \textit{can a single level diffusion LM use its existing noise level to decide when to plan and when to polish?}

Our idea is to treat the noise level as a granularity cue.
As illustrated in Figure~\ref{fig:fig1}, high-noise denoising should operate on more coherent evidence and make coarse meaning commitments, while lower-noise steps should return to token-level refinement.
If training exposure and inference commitment follow this same schedule, the model can form plan-like structure before local wording without an explicit planner or hierarchy.
Specifically, we propose \emph{Noise Dependent Granularity Control} (NDGC).

During training, NDGC changes visible evidence via noise-dependent span masking, exposing span-correlated evidence while preserving the same marginal corruption rate. As noise decreases, this schedule transitions smoothly back to tokenwise exposure. 
During inference, NDGC mirrors the same granularity schedule: high-noise steps commit groups by aggregate confidence, while low-noise steps commit individual tokens by token confidence.
This alignment moves denoising from coarse organization to local wording.

We evaluate NDGC by asking when structure appears during denoising.
Our main test is a controlled topic skeleton benchmark, where each target follows an ordered hidden topic skeleton.
This lets us measure final skeleton recovery and early skeleton formation.
In our high-headroom BD3LM setting, uniform tokenwise denoising tends to form structure late or repeat.
NDGC improves early structure, ordered recovery, and output health on the main backbone.
Ablations show why alignment matters.
Structured exposure alone creates a mismatch between training and inference.
Forcing token commitment at high noise delays ordered skeleton formation even when final topic recovery is high.
Additional backbones and WritingPrompts give bounded support beyond the controlled task.

{\bf Contributions.} The key contributions include:
\begin{itemize}
    \item We identify fixed token granularity across noise levels as a limitation of tokenwise discrete diffusion denoising for generation that needs planning.
    \item We introduce \emph{Noise Dependent Granularity Control} (NDGC), which couples structured exposure during training with matched commitment during inference, without adding an explicit planner or hierarchy.
    \item We provide topic skeleton diagnostics and ablation evidence showing that NDGC shifts structure formation earlier, improves ordered recovery, and supports healthier outputs.
\end{itemize}

\section{Related Work}

Discrete diffusion language models generate text by denoising corrupted sequences, with work on state spaces, parameterizations, training, and sampling \citep{austin2021structured,lou2023discrete,nie2025scaling}. Concurrent work by \citet{sahoo2024simple} and \citet{shi2024simplified} independently derived simplified masked diffusion formulations and variational objectives and demonstrated strong language modeling results; \citet{shi2024simplified} additionally generalized the framework to state-dependent masking schedules. Iterative refinement and insertion models likewise show that reveal order matters \citep{lee2018deterministic,gu2019levenshtein,stern2019insertion}. Still, the corruption and commitment unit is usually token level or fixed by schedule. 

Long form generation often separates content planning from wording with outlines, sketches, or plans \citep{fan2018hierarchical,yao2019plan,fan2019strategies,yang2022re3}. Block based diffusion and story systems make this split explicit through blocks, intermediate structure, or extra stages \citep{arriola2025block,DBLP:conf/acl/LeiGHZZPL024,DBLP:conf/eacl/XieR24,DBLP:conf/iclr/HuotAPJCL25}. These methods add planners, latents, or separate pipelines. NDGC asks whether a single level conditional diffusion model can use its own noise level to plan early and polish late, while closely related work includes outline guided and recursive planning methods \citep{DBLP:conf/naacl/WangHLWLHT25,DBLP:conf/naacl/LeeKSKK25,xiong2025outliningheterogeneousrecursiveplanning}.

Span corruption in denoising pretraining shows that contiguous missing regions can support broader context use than independent token masking \citep{DBLP:conf/acl/ZhuKW025,DBLP:conf/acl/LeeKYY25,DBLP:conf/coling/AsadaM25a,DBLP:conf/iclr/LiuNCSXJG25}. These objectives are usually static corruptions, not reverse generation steps indexed by noise. Adaptive unmasking changes inference commitment, but is not always matched to training evidence \citep{DBLP:conf/icml/ZhangZ0TOSJ25,DBLP:conf/icml/RutteFDOS025,DBLP:conf/iclr/WangUHWLJLR0B25}. NDGC links exposure and commitment across the denoising trajectory: high noise uses coherent groups, while low noise returns to token refinement.

\section{Background: Discrete Denoising Diffusion Probabilistic Models}
\label{sec:background_d3pm}

\noindent\textbf{Notation.}
Let \(c\) be the conditioning context and let
\(y=(y_1,\ldots,y_L)\) be the target answer. The answer positions are
\(\mathcal{A}=\{1,\ldots,L\}\). At diffusion step \(t\),
\(x_t=(x_{t,1},\ldots,x_{t,L})\) is the corrupted answer, and
\(r(t)\in[0,1]\) is the masking ratio. Larger \(r(t)\) means higher noise.
The model distribution is \(p_\theta(\cdot\mid x_t,c,t)\). The coefficient
\(s(t)\in[0,1]\) gives the granularity level, and \(B(t)\) is the target group
length in tokens.

\noindent\textbf{Diffusion setup.}
We use the standard discrete denoising diffusion setup. A Markov forward
process corrupts a clean token sequence, and a learned reverse process
denoises it step by step
\citep{austin2021structured,lou2023discrete,sahoo2024simple,shi2024simplified}. Let
\(x_0=(x_0^1,\ldots,x_0^L)\) be a clean sequence, where each token belongs to
a vocabulary \(\mathcal{V}\). Given \(T\) diffusion steps, we write
\(t_j=j/T\).

The forward process specifies how noise is added. It gradually corrupts the
clean sequence through token transition matrices:
\begin{equation}
\begin{aligned}
q(x_{t_1:T}\mid x_0)
&=
\prod_{j=1}^{T}
q(x_{t_j}\mid x_{t_{j\ndminus 1}}),\\
q(x_{t_j}^{\ell}\mid x_{t_{j\ndminus 1}}^{\ell})
&=
\operatorname{Cat}
\left(
x_{t_j}^{\ell}
\mid
Q_j^{\top}x_{t_{j\ndminus 1}}^{\ell}
\right).
\end{aligned}
\label{eq:d3pm_forward_short}
\end{equation}
Here \(Q_j\) is a token transition matrix. Different choices of \(Q_j\)
define different corruption processes, including absorbing mask corruption.

The reverse process specifies how the model removes noise. Given a noisy
state, the model predicts a less corrupted state by combining clean token
beliefs with the known forward posterior:
\begin{equation}
\begin{aligned}
p_\theta(x_{t_{j\ndminus 1}}\mid x_{t_j})
&=
\prod_{\ell=1}^{L}
p_\theta(x_{t_{j\ndminus 1}}^{\ell}\mid x_{t_j}),\\
p_\theta(x_{t_{j\ndminus 1}}^{\ell}\mid x_{t_j})
&=
\sum_{\hat{x}^{\ell}\in\mathcal{V}}
q(x_{t_{j\ndminus 1}}^{\ell}
\mid x_{t_j}^{\ell},\hat{x}^{\ell})\\
&\quad\cdot
p_\theta(\hat{x}^{\ell}\mid x_{t_j},t_j).
\end{aligned}
\label{eq:d3pm_reverse_short}
\end{equation}
In masked text diffusion, the neural model is commonly trained to predict
clean tokens from corrupted states.

\noindent\textbf{Conditional masked diffusion.}
Building on this diffusion setup, this paper focuses on the conditional
masked case, where a visible context is kept fixed and only the target answer
is denoised. 
The context stays
visible, and only answer positions are corrupted. With masking ratio
\(r(t)\in[0,1]\), the standard tokenwise corruption is
\begin{equation}
\begin{gathered}
x_{t,i} =
\begin{cases}
\texttt{[MASK]}, & m_i=1,\\
y_i, & m_i=0,
\end{cases}
\end{gathered}
\label{eq:background_tokenwise_mask_short}
\end{equation}
where $m_i\sim\operatorname{Bernoulli}(r(t))$.
The model predicts \(p_\theta(\cdot\mid x_t,c,t)\) over answer tokens. In this
standard conditional setting, \(r(t)\) controls how many answer tokens are
hidden, while both corruption and commitment remain token level at every noise
level. This leaves the denoising granularity fixed. The next section introduces a method that makes granularity noise dependent while keeping the same diffusion backbone.

\section{Method: Noise Dependent Granularity Control}
\label{sec:method}

\begin{figure}[t]
\centering
\includegraphics[width=0.95\linewidth]{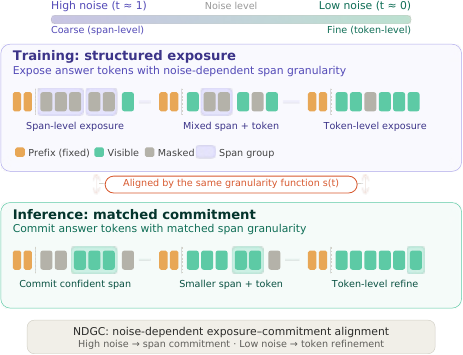}
\caption{
\textbf{Noise-Dependent Granularity Control (NDGC).}
NDGC uses the noise level to choose the denoising granularity. At high noise,
training exposes coherent token groups and inference commits high-confidence
groups; at low noise, both return to token-level refinement. The same schedule
$s(t)$ determines the group size $B(t)$ in both training and inference, aligning
what the model sees during training with what the sampler commits during
generation. This creates coarse-to-fine denoising without an explicit planner
or hierarchy.
}
\label{fig:granularity_axis}
\end{figure}

We propose Noise-Dependent Granularity Control (NDGC), a single-level diffusion
method that uses noise level as a granularity signal. This section first describes structured exposure during training in Section~\ref{sec:structured_exposure}, then
describes matched commitment during inference in Section~\ref{sec:structured_commitment}.


\subsection{Structured exposure during training.}
\label{sec:structured_exposure}

\noindent\textbf{Structured exposure.}
Standard tokenwise corruption treats each answer token independently at every
noise level. This is undesirable at high noise: the few visible tokens are often
scattered, so the model receives weak local clues rather than coherent evidence.
Structured exposure keeps the same expected masking ratio \(r(t)\), but changes
the correlation pattern of the corruption. At high noise, nearby answer tokens
are masked or exposed in groups; at low noise, the process returns to tokenwise
masking.

We define the noise-dependent granularity coefficient
\begin{equation}
s(t)
=
\operatorname{clip}_{[0,1]}
\left(
\frac{r(t)\ndminus r_{\mathrm{low}}}
     {r_{\mathrm{high}}\ndminus r_{\mathrm{low}}}
\right).
\label{eq:granularity_coefficient}
\end{equation}
Thus \(s(t)\) is large when the noise level is high and small when the noise
level is low. Given a nominal group length \(B_{\max}\), we set
\begin{equation}
B(t)
=
\max\!\left(
1,
\operatorname{round}\!\left(1+s(t)(B_{\max}\ndminus 1)\right)
\right).
\label{eq:span_length}
\end{equation}
The answer region is then partitioned into contiguous groups, i.e., consecutive
blocks of answer tokens, using target length \(B(t)\). These groups are not
learned semantic spans; they are a simple token-order partition used to control
the granularity of corruption.
Importantly, these groups do not require access
to semantic boundaries; they are simple consecutive blocks of answer tokens
whose size is controlled by the noise level.

Structured exposure samples masks over these groups while preserving the
marginal masking probability of each token. Therefore, each answer token is
still masked with probability \(r(t)\) in expectation, but high-noise examples
contain span-correlated visible evidence instead of isolated token remnants.
The exact partition rule and marginal preservation are given in
Appendix~\ref{app:ndgc_details}.

\noindent\textbf{Training objective.}
NDGC uses the same token prediction loss as the baseline. It only changes how
the corrupted state \(x_t\) is sampled. Let \(\lambda(t)\) be the diffusion
loss weight of the backbone. Define
\begin{equation}
\ell_i(\theta)
=
\operatorname{CE}\bigl(
p_\theta(\cdot\mid x_t,c,t), y_i
\bigr).
\end{equation}
The training objective is
\begin{multline}
\mathcal{L}_{\mathrm{NDGC}}(\theta)
=
\mathbb{E}_{(c,y),t,x_t}
\left[
\sum_{i\in \mathcal{A}}\lambda(t)\ell_i(\theta)
\right].
\label{eq:ndgc_objective}
\end{multline}
Thus NDGC adds no prediction head, latent variable, or auxiliary loss. It only
replaces the tokenwise exposure distribution with a noise-dependent structured
one, making it easy to apply to existing masked diffusion backbones with minimal
architectural changes.

\subsection{Structured commitment during inference.}
\label{sec:structured_commitment}

Structured commitment is the inference-side counterpart of structured exposure.
If high-noise training presents evidence at the group level, then high-noise
sampling should also make decisions at the group level. Otherwise, the sampler
would return to isolated token commitments even though the model was trained to
use coarser evidence. NDGC therefore uses the same \(s(t)\) and \(B(t)\) during
sampling: high noise commits confident groups, while low noise returns to token
refinement.

At each denoising step, we partition the answer positions into groups
\(G_j(t)\) using \(B(t)\), and consider only the positions that remain masked:
\begin{equation}
U_j(t)=\{i\in G_j(t): x_{t,i}=\texttt{[MASK]}\}.
\label{eq:still_masked_group}
\end{equation}
For each still-masked token, we compute the confidence of the best non-mask
prediction:
\begin{equation}
\gamma_i(t)
=
\max_{v\in\mathcal{V}\setminus\{\texttt{[MASK]}\}}
p_\theta(v\mid x_t,c,t).
\label{eq:token_confidence}
\end{equation}
The confidence of a group is the average confidence over its still-masked
positions:
\begin{equation}
\Gamma_j(t)
=
\frac{1}{|U_j(t)|}
\sum_{i\in U_j(t)} \gamma_i(t).
\label{eq:span_confidence}
\end{equation}

The sampler ranks groups by \(\Gamma_j(t)\), selects high-confidence groups
under the step budget, and samples their tokens from the reverse distribution.
Unselected groups remain masked, while previously revealed tokens stay fixed.
When \(B(t)=1\), each group contains one token, so the same rule reduces to
standard token-level refinement. The budget rule is given in
Appendix~\ref{app:ndgc_details}.

Together, we implement the exposure--commitment alignment
shown in Figure~\ref{fig:granularity_axis}: the noise level determines both what scale of
evidence the model sees during training and what scale of decisions the sampler
makes during inference.

\section{Experimental Setup}
\label{sec:setup}
We test whether noise-dependent exposure--commitment alignment changes denoising dynamics, not only final scores. We denote structured exposure as SE, structured commitment as SC, and budget matched tokenwise inference as BM-TOK. Table~\ref{tab:systems} gives the matched controls.

\begin{figure}[t]
\centering
\includegraphics[width=.78\linewidth]{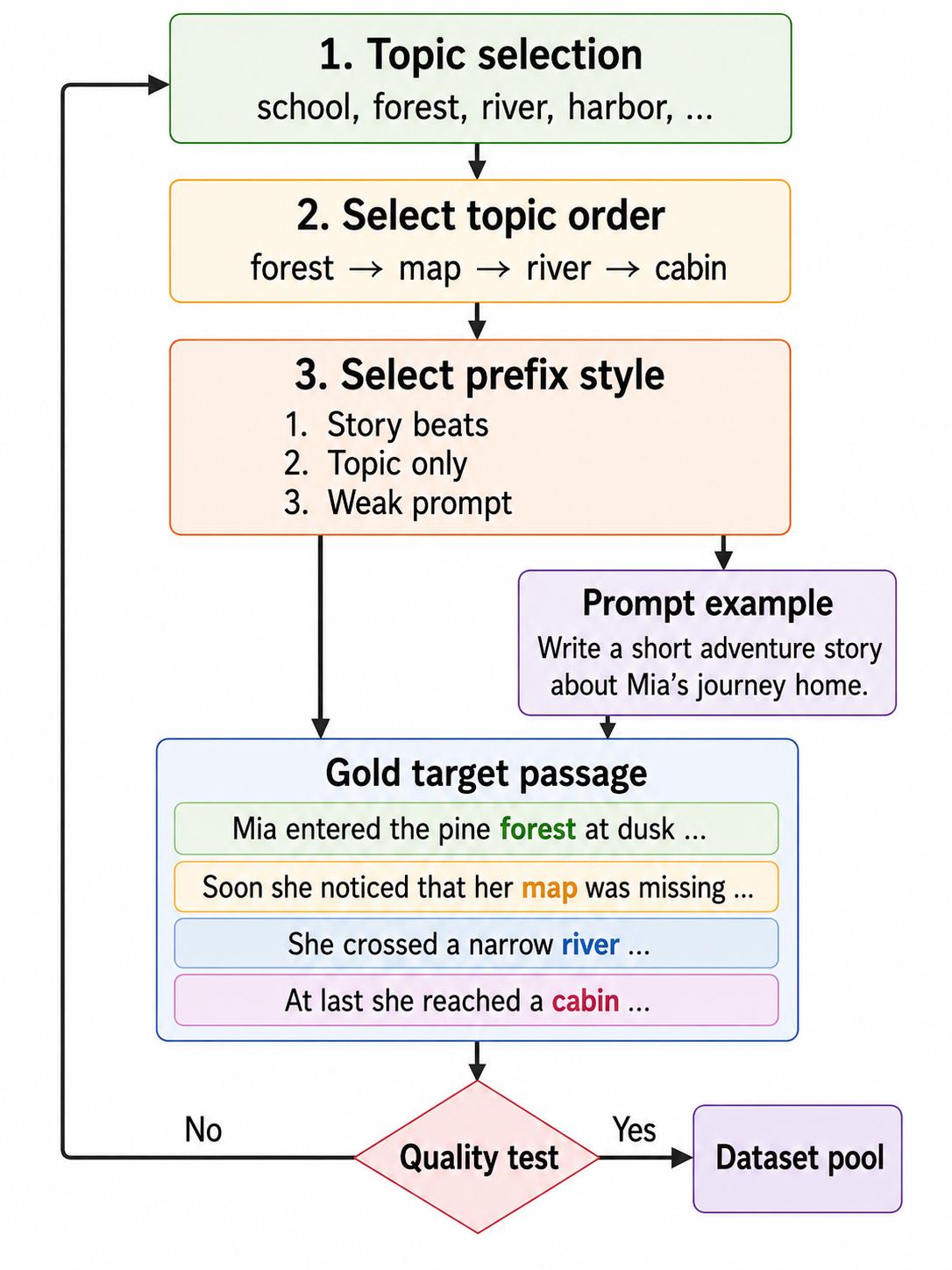}
\caption{
\textbf{\textsc{Synthetic-V4} pipeline.}
Topics are sampled, ordered into a latent skeleton, exposed through one prompt view, and realized as topic-aligned discourse spans. Only examples passing automatic quality checks enter the dataset pool.}
\label{fig:synthetic_v4_pipeline}
\end{figure}

\noindent\textbf{Data.}
Synthetic-V4 is our primary controlled benchmark because it makes planning-like behavior directly measurable. As shown in Figure~\ref{fig:synthetic_v4_pipeline}, each example is generated by sampling a topic set, ordering the topics into a latent topic skeleton, selecting a prompt-observability regime, and realizing the skeleton as a target passage with topic-aligned discourse spans. We train on 50{,}000 examples and evaluate on 500 held-out examples. The model observes only the prompt; gold topic order, span boundaries, and transition labels are used only by the evaluator. Sequences are capped at 1{,}024 tokens, with prefixes up to 192 tokens. 
WritingPrompts \citep{fan2018hierarchical} is a naturalistic BD3LM check on held-out prompts with one reference continuation per prompt.
Since it has no gold skeleton, we report only output health, repetition, and reference-aligned segment similarity. Dataset details are in Appendix~\ref{app:data_synthetic}.

\noindent\textbf{Systems and implementation.}
We adapt BD3LM \citep{arriola2025block},
MDLM \citep{sahoo2024simple}, and SEDD \citep{lou2023discrete} to conditional generation. BD3LM is used for the full control matrix and denoising trajectories; MDLM and SEDD provide \textsc{Synthetic-V4} backbone checks. Within each backbone, all systems share the same data, evaluation set, sampling steps, nucleus setting, and answer length limits. Unless varied, structured variants use \(B_{\max}=16\), \(r_{\mathrm{low}}=0.3\), and \(r_{\mathrm{high}}=0.7\); Section~6.1 reports the \(B_{\max}\) sensitivity check. Final inference uses the best validation checkpoint. Runs used 2 NVIDIA H100 GPUs on DeltaAI GH200 nodes.

\begin{table*}[t]
\centering
\small
\setlength{\tabcolsep}{3.5pt}
\renewcommand{\arraystretch}{1.08}
\caption{Experimental systems and the mechanistic question addressed by each intervention.}
\label{tab:systems}
\begin{tabularx}{\textwidth}{p{0.20\textwidth} p{0.18\textwidth} p{0.22\textwidth} X}
\toprule
System & Training exposure & Inference commitment & Mechanistic question \\
\midrule
Baseline
& Uniform tokenwise
& Uniform tokenwise
& Does standard tokenwise denoising lack an explicit planning-granularity signal? \\

Structured commitment only
& Uniform tokenwise
& Structured commitment
& Is the effect merely a decoder-side heuristic? \\

Structured exposure only
& Structured exposure
& Uniform tokenwise
& Is structured exposure alone sufficient, or does it create a train--inference granularity mismatch? \\

Tokenwise early
& Structured exposure
& Tokenwise at high noise
& Does forcing token-level high-noise commitment delay early skeleton formation? \\

Budget-matched tokenwise
& Structured exposure
& Tokenwise with matched reveal budget
& Is the gain due to span-level commitment rather than reveal pacing alone? \\

NDGC (ours)
& Structured exposure
& Matched structured commitment
& Does exposure--commitment alignment induce coarse-to-fine denoising? \\
\bottomrule
\end{tabularx}
\end{table*}

\textbf{Metrics.}
On \textsc{Synthetic-V4}, the evaluator extracts a predicted topic sequence from each output and compares it with the gold ordered skeleton. Final coverage is the fraction of gold topics that appear in the predicted sequence. Order is the normalized longest common subsequence score between the predicted and gold topic sequences. Exact is one only when the complete predicted sequence matches the gold sequence. Early recovery is final coverage averaged over partial denoising snapshots at fixed reveal fractions. Early order avg measures how well the model has already arranged the recovered topics in the correct gold order at those partial snapshots during early denoising.

Bad output is the rate of generations that trigger any severe validity or degeneration flag. It is used as an output health check, not as a planning score. On WritingPrompts, we report RASC best, a diagnostic we introduce for reference-aligned segment similarity. It splits generated and reference stories into macro segments, embeds each segment with a fixed sentence encoder, and averages the best reference match for each generated segment. Full extraction rules, flag thresholds, repetition checks, and RASC best details are in Appendix~\ref{app:metric-details}.

\section{Experimental Results}
\begin{table*}[t]
\centering
\small
\caption{
\textbf{BD3LM results on Synthetic-V4.} Values are two seed means in percentages.
Final metrics measure topic coverage, order, and exact skeleton recovery.
Early metrics average recovery over partial denoising snapshots.
Higher is better except bad output.}
\label{tab:bd3lm_synthetic_v4}
\resizebox{\textwidth}{!}{
\begin{tabular}{lcccccc}
\toprule
System
  & \makecell{Final topic\\coverage $\uparrow$}
  & \makecell{Order\\acc. $\uparrow$}
  & \makecell{Exact\\skeleton $\uparrow$}
  & \makecell{Early recovery\\avg. $\uparrow$}
  & \makecell{Early order\\avg. $\uparrow$}
  & \makecell{Bad\\output $\downarrow$} \\
\midrule
BD3LM (baseline)
  & $27.8\%$ & $26.6\%$ & $2.0\%$ & $28.7\%$ & $27.5\%$ & $99.7\%$ \\
SC only
  & $54.5\%$ & $46.6\%$ & $0.3\%$ & $56.9\%$ & $47.9\%$ & $89.0\%$ \\
SE only
  & $58.6\%$ & $52.4\%$ & $0.1\%$ & $51.7\%$ & $46.1\%$ & $19.8\%$ \\
Tokenwise early
  & $82.7\%$ & $78.8\%$ & $46.7\%$ & $57.5\%$ & $53.6\%$ & $8.3\%$ \\
BM-TOK
  & $82.71\%$ & $78.4\%$ & $44.4\%$ & $57.9\%$ & $53.9\%$ & $7.0\%$ \\
\midrule
NDGC ($B_{\max}=8$)
 & $81.1 \%$ & $76.6 \%$ & $39.0 \%$ & $77.0 \%$ &$72.9 \%$ &  \textbf{2.4\%} \\
NDGC ($B_{\max}=32$)
 & $82.7 \%$ & $78.9 \%$ & $47.0 \%$ & $74.5 \%$ &$71.1 \%$ & $5.0 \%$ \\
NDGC ($B_{\max}=16$)
  & \textbf{84.2\%} & \textbf{79.8\%} & \textbf{50.4\%} & \textbf{79.1\%} & \textbf{77.0\%} &3.7\% \\
  
\bottomrule
\end{tabular}
}
\end{table*}

\subsection{Controlled Result and Mechanism Analysis: BD3LM Synthetic-V4}
\label{sec:bd3lm_synthetic_v4}

\begin{figure*}[t]
\centering
\begin{minipage}{0.53\textwidth}
\centering
\includegraphics[width=\linewidth]{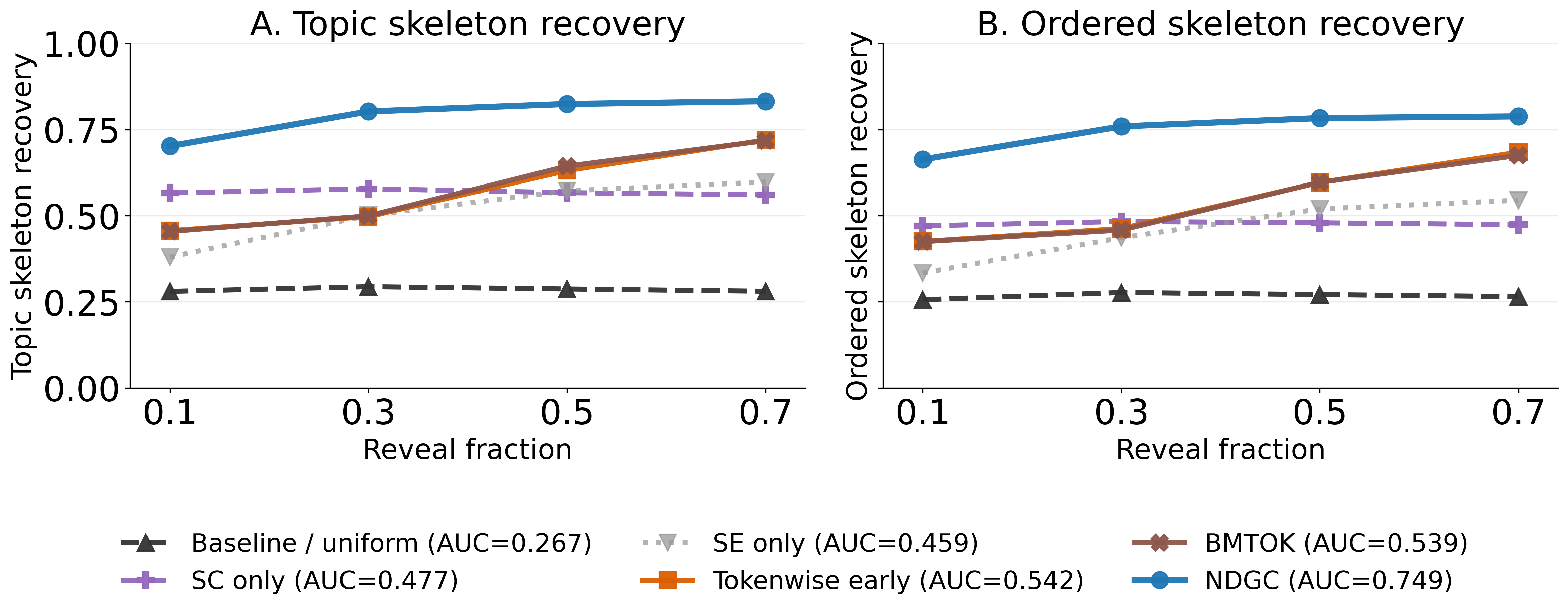}
\end{minipage}
\hfill
\begin{minipage}{0.42\textwidth}
\centering
\includegraphics[width=\linewidth]{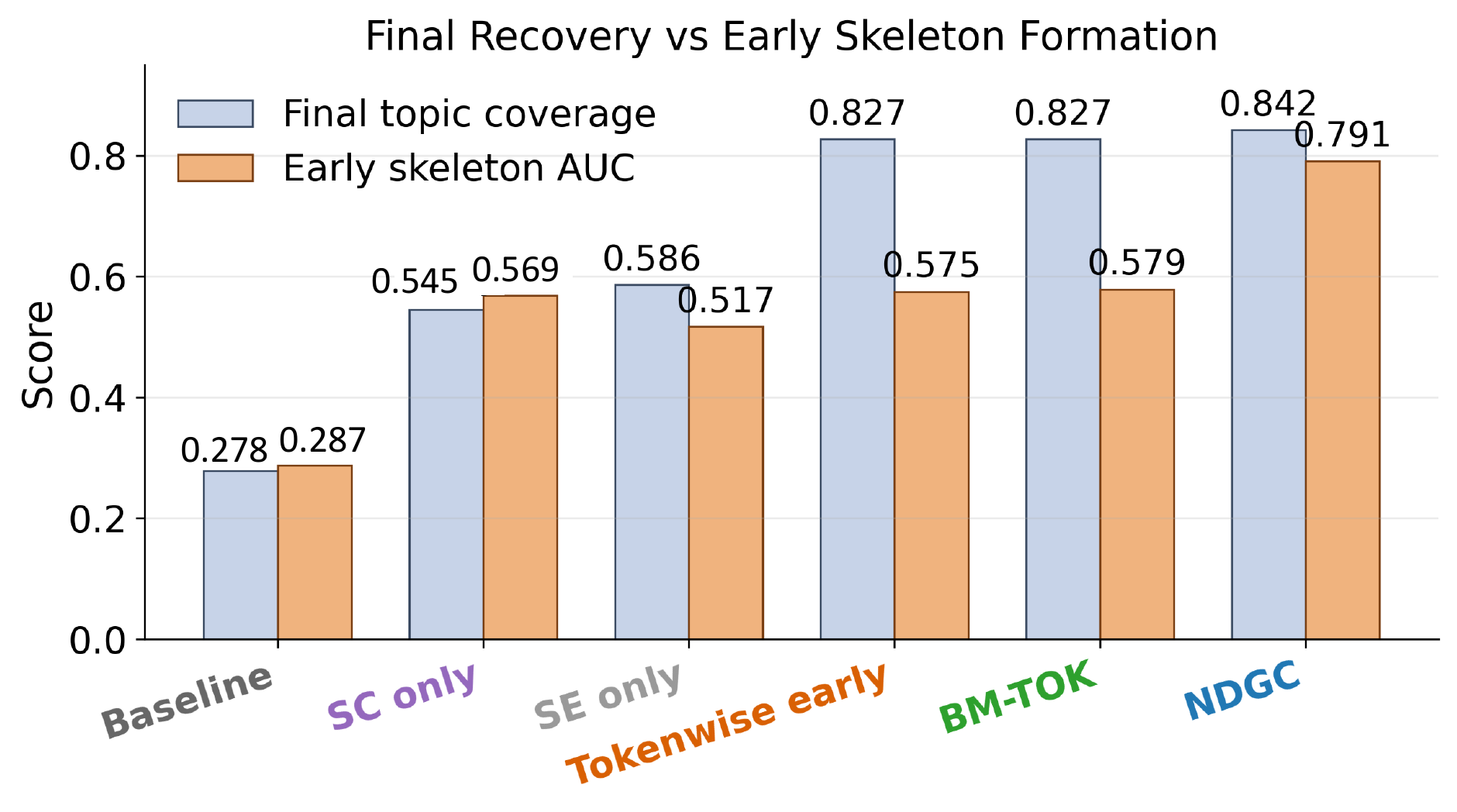}
\end{minipage}
\caption{
\textbf{Final recovery and early skeleton formation on BD3LM Synthetic-V4.}
The left plot gives topic and ordered skeleton recovery trajectories over
reveal fractions. The right plot compares final topic coverage with early
skeleton AUC. NDGC forms recoverable skeleton structure earlier than the
baseline and partial controls. Tokenwise early reaches final coverage close to
NDGC but stays much lower in early recovery, which separates late topic repair
from early skeleton formation.
}
\label{fig:bd3lm_trajectory_summary}
\end{figure*}

\begin{table*}[t]
\centering
\small
\caption{
\textbf{Compact bad output decomposition on BD3LM Synthetic-V4.} All values are
reported as percentages. The baseline failure is driven by repetition and
diversity rather than validity artifacts.
}
\label{tab:bd3lm_bad_output_decomp}

\begin{tabular}{lcccccc}
\toprule
Trigger group
  & Base
  & SC
  & SE
  & Tok early
  & BM-TOK
  & NDGC \\
\midrule
Non repetition validity
  & $0.0\%$ & $1.2\%$ & $0.0\%$ & $0.0\%$ & $0.0\%$ & $0.0\%$ \\
Repetition and diversity
  & $99.7\%$ & $89.0\%$ & $19.8\%$ & $8.3\%$ & $7.0\%$ & $3.7\%$ \\
\texttt{Bad\_output}
  & $99.7\%$ & $89.0\%$ & $19.8\%$ & $8.3\%$ & $7.0\%$ & $3.7\%$ \\
\bottomrule
\end{tabular}

\end{table*}

\noindent\textbf{NDGC gives the best balance.}
Synthetic-V4 is our most controlled setting because each target is built from an ordered latent topic skeleton, so we can measure both final skeleton recovery and when the skeleton first appears during denoising. This makes conditional BD3LM a test of early ordered semantic commitment, not a claim that BD3LM is a weak unconditional backbone. Table 2 should be read as a high-headroom BD3LM mechanism test, not evidence that tokenwise diffusion universally collapses. Table~\ref{tab:bd3lm_synthetic_v4} shows that uniform tokenwise BD3LM fails this conditional long-form test: final topic coverage is only $27.8\%$, order accuracy is $26.6\%$, exact skeleton match is $2.0\%$, Early recovery avg. is $28.7\%$, Early order avg. is $27.5\%$, and bad output is $99.7\%$. NDGC with $B_{\max}=16$ gives the strongest overall result. It raises final coverage to $84.2\%$, order accuracy to $79.8\%$, exact skeleton match to $50.4\%$, Early recovery avg. to $79.1\%$, and Early order avg. to $77.0\%$, while reducing bad output to $3.7\%$. The scale check also matters: $B_{\max}=8$ is safest on bad output at $2.4\%$ but weaker on skeleton recovery, while $B_{\max}=32$ keeps strong final recovery but loses some early structure. We therefore use $B_{\max}=16$ as the default.

\noindent\textbf{Both sides are needed.}
The partial controls show that neither side of NDGC explains the result by itself. \textsc{SC only} applies structured commitment to a uniformly trained model. It raises final coverage to $54.5\%$ and Early recovery avg. to $56.9\%$, but bad output remains $89.0\%$ and exact skeleton match is only $0.3\%$, so the improvement is not merely a decoder-side heuristic. \textsc{SE only} trains with structured exposure but keeps tokenwise inference. It improves coverage to $58.6\%$ and lowers bad output to $19.8\%$, but exact match is still $0.1\%$ and Early recovery avg. is $51.7\%$. Thus structured exposure without matched commitment leaves a train--inference granularity mismatch.

\noindent\textbf{Tokenwise commitment is late.}
\textsc{Tokenwise early} and \textsc{BM-TOK} use the structured trained model, and \textsc{BM-TOK} also matches the NDGC reveal budget, but both still commit tokens independently. These controls reach high final coverage, $82.7\%$ and $82.71\%$, with exact matches of $46.7\%$ and $44.4\%$. However, their Early recovery avg. scores are only $57.5\%$ and $57.9\%$, and their Early order avg. scores are $53.6\%$ and $53.9\%$, far below NDGC at $79.1\%$ and $77.0\%$. Figure~\ref{fig:bd3lm_trajectory_summary} shows the same separation visually: NDGC rises earlier in both topic recovery and ordered recovery, while the tokenwise controls stay close only in final coverage. This rules out reveal budget alone and shows why final topic scores are not enough.

\begin{table*}[t]
\centering
\small
\setlength{\tabcolsep}{4pt}
\renewcommand{\arraystretch}{1.12}
\caption{
\textbf{Qualitative topic coverage example from BD3LM Synthetic-V4.}
The baseline repeats the first topic, while NDGC covers the requested
ordered topic sequence with topic specific evidence.
}
\label{tab:qual_topic_case}
\begin{tabularx}{\textwidth}{lYY}
\toprule
\textbf{Gold topic} & \textbf{BD3LM baseline} & \textbf{NDGC} \\
\midrule
\textbf{Topic trace}
&
\textbf{newsroom $\rightarrow$ newsroom $\rightarrow$ newsroom $\rightarrow$ newsroom $\rightarrow$ newsroom}
&
\textbf{newsroom $\rightarrow$ space $\rightarrow$ school $\rightarrow$ desert $\rightarrow$ museum}
\\
\midrule
\textsc{newsroom}
&
Recovered, but repeated. The output keeps returning to
``fact checker noticed the fact'' and related newsroom phrases.
&
``The producer brought the camera card back to the news desk, where the
anchor had already marked the quote line ...''
\\

\textsc{space}
&
Not recovered as a separate section. The output continues newsroom and
fact checking language.
&
``The controller asked the engineer to compare the ping with the
thruster valve before the first late radio signal became harder to
contain...''
\\

\textsc{school}
&
Not recovered as a separate section. The output continues newsroom and
fact checking language.
&
``The mentor asked the teacher to compare the answer with the project
board near the science room before ...''
\\

\textsc{desert}
&
Not recovered as a separate section. The output continues newsroom and
fact checking language.
&
``The caravan leader protected the dune until the guide note pointed to
the dry well...''
\\

\textsc{museum}
&
Not recovered as a separate section. The output continues newsroom and
fact checking language.
&
``The docent kept the display label near the storage wing until the
pedestal made the misplaced artifact clearer...''
\\
\bottomrule
\end{tabularx}
\end{table*}

\noindent\textbf{Repetition drives baseline failure.}
Table~\ref{tab:bd3lm_bad_output_decomp} shows that the uniform tokenwise baseline has $0.0\%$ non-repetition validity failure, so it is not mainly empty text, token leakage, extreme length, or control-character artifacts. Instead, the failure comes from repetition and diversity: this trigger group is $99.7\%$ for the baseline and only $3.7\%$ for NDGC. This matters for interpreting Table~\ref{tab:bd3lm_synthetic_v4}. The result should be read as a conditional long-form failure of uniform tokenwise denoising, where the model tends to repeat and make local repairs when it must make ordered semantic commitments. Table~\ref{tab:qual_topic_case} gives the same failure in one example: the baseline repeats the first topic, while NDGC moves through the requested topic sequence.

\noindent\textbf{Health is not the mechanism.}
The health improvements interact with, but do not replace, the early-skeleton evidence. \textsc{SC only} remains unhealthy, \textsc{SE only} improves health but does not produce strong exact or early ordered recovery, and the tokenwise controls reduce bad output to $8.3\%$ and $7.0\%$ while remaining much weaker than NDGC in early recovery. Therefore the main effect is not simply repetition control. The decisive pattern is that NDGC improves health while also shifting recoverable ordered structure earlier in the denoising trajectory.

\subsection{Cross Backbone Evidence: MDLM and SEDD}
\label{sec:cross_backbone}

Section~\ref{sec:bd3lm_synthetic_v4} shows that NDGC changes BD3LM denoising
by forming the ordered skeleton earlier, not just by repairing topics at the
end. We next check whether this reading survives beyond the high headroom BD3LM
case. MDLM and SEDD have healthy uniform baselines on Synthetic-V4, so there
is little degeneration to remove. This makes the comparison a stricter test of
whether the same noise dependent granularity idea still helps.

\begin{table*}[t]
\centering
\normalsize
\caption{
\textbf{Cross backbone Synthetic-V4 results under healthy uniform baselines.} All values
are two seed means and are reported as percentages. Higher is better for
coverage, order accuracy, exact skeleton, early recovery, and early order.
Lower is better for bad output.
}
\label{tab:cross_backbone}
\setlength{\tabcolsep}{5pt}
\resizebox{\textwidth}{!}{
\begin{tabular}{lrrrrrr}
\toprule
System
  & \makecell{Final topic\\coverage $\uparrow$}
  & \makecell{Order\\acc. $\uparrow$}
  & \makecell{Exact\\skeleton $\uparrow$}
  & \makecell{Early recovery\\avg. $\uparrow$}
  & \makecell{Early order\\avg. $\uparrow$}
  & \makecell{Bad\\output $\downarrow$} \\
\midrule
MDLM
  & $62.4\%$ & $45.8\%$ & $0.2\%$ & $59.8\%$ & $43.7\%$ & $0.00\%$ \\
MDLM + \textsc{Tokenwise early}
  & $64.3\%$ & $52.5\%$ & $3.0\%$ & $52.4\%$ & $52.5\%$ & $3.2\%$ \\
MDLM + \textsc{BM-TOK}
  & $63.1\%$ & $43.3\%$ & $3.0\%$ & $54.3\%$ & $43.3\%$ & $3.1\%$ \\
MDLM + \textsc{NDGC}
  & \textbf{66.4\%} & \textbf{55.0\%} & \textbf{7.0\%} & \textbf{63.0\%} & \textbf{52.5\%} & \textbf{0.0\%} \\
\midrule
SEDD
  & $61.9\%$ & $46.2\%$ & $0.0\%$ & $59.6\%$ & $45.1\%$ & $0.0\%$ \\
SEDD + \textsc{Tokenwise early}
  & $58.7\%$ & $45.7\%$ & $0.4\%$ & $49.6\%$ & $40.1\%$ & $2.0\%$ \\
SEDD + \textsc{BM-TOK}
  & $62.4\%$ & $47.5\%$ & $0.3\%$ & $50.6\%$ & $44.5\%$ & $17.2\%$ \\
SEDD + \textsc{NDGC}
      & \textbf{62.6\%} & \textbf{48.4\%} & \textbf{0.4\%} & \textbf{59.9\%} & \textbf{46.2\%} & \textbf{0.0\%} \\
\bottomrule
\end{tabular}
}
\end{table*}

\noindent\textbf{MDLM shows a clean gain.} Table~\ref{tab:cross_backbone} shows a clear MDLM gain even when the baseline is already output healthy. NDGC improves the main final and early skeleton scores while keeping bad output at zero. The tokenwise controls recover some topics, but they do not give the same early recovery pattern. This matches the logic of Section~\ref{sec:bd3lm_synthetic_v4}: final topic repair is not the same as forming the skeleton early. \noindent\textbf{SEDD gives bounded support.} SEDD gives a smaller signal, but it still moves in the right direction for the main skeleton metrics. The health result is also useful. BM-TOK can improve some final scores, but it raises bad output sharply, while NDGC keeps the healthy baseline intact. These results support a narrow conclusion. The gain size is backbone dependent, but matching the granularity of training evidence and inference commitment is the safest choice among the tested systems. This cross backbone check also helps rule out the simple reading that NDGC only fixes a BD3LM specific degeneration rather than broader behavior.

\subsection{WritingPrompts Naturalistic Corroboration}
\label{sec:wp}
\begin{table}[t]
\centering
\small
\caption{
\textbf{WritingPrompts results on BD3LM.} Values are percentages. Lower is better for bad output and Rep 4 gram, and higher is better for RASC best. This is a natural story generation check, not direct skeleton evidence. RASC-best is cosine similarity averaged over generated segments and multiplied by 100.
}
\label{tab:writingprompts}
\resizebox{\columnwidth}{!}{%
\begin{tabular}{lccc}
\toprule
System 
& Bad output $\downarrow$ 
& Rep 4 gram $\downarrow$ 
& RASC best $\uparrow$ \\
\midrule
Baseline 
& $22.8\%$ 
& $3.10\%$ 
& $31.3\%$ \\
SC only 
& $30.4\%$ 
& $10.3\%$
& $31.6\%$ \\
SE only 
& $91.9\%$ 
& $79.2\%$ 
& $17.4\%$ \\
Tokenwise early 
& $30.1\%$ 
& $0.95\%$ 
& $27.6\%$ \\
NDGC
& $10.9\%$ 
& $1.00\%$ 
& $32.8\%$ \\
\bottomrule
\end{tabular}%
}
\end{table}

\noindent\textbf{NDGC improves the balance.}
Table~\ref{tab:writingprompts} reports a natural conditional story generation check on WritingPrompts with BD3LM. WritingPrompts has no gold latent topic skeleton, so we use it for output health, local repetition, and reference aligned segment similarity, not for direct skeleton recovery. This is a weaker but useful check, because the evaluator cannot observe a hidden plan.
NDGC improves all three measures. Bad output drops from $22.8\%$ to $10.9\%$, Rep 4 gram drops from $3.10\%$ to $1.00\%$, and RASC best rises from $31.3\%$ to $32.8\%$. The controls give the same reading as Synthetic-V4. \textsc{SE only} is unstable, with $91.9\%$ bad output and $79.2\%$ Rep 4 gram. \textsc{SC only} gives a small RASC best gain, but worsens health and repetition. \textsc{Tokenwise early} keeps Rep 4 gram low, but raises bad output and lowers RASC best. Thus low local repetition alone is not enough. Only NDGC gives the best balance across health, repetition, and segment similarity. The pattern therefore matches the controlled task at the level of output behavior. We read this result as natural support for granularity control, not as direct evidence of latent skeleton recovery.

\section{Conclusion}
NDGC aligns exposure and commitment. High noise uses coherent groups for coarse organization, while low noise refines tokens. Topic skeleton diagnostics and ablations show earlier skeleton formation beyond reveal pacing, late repair, or repetition control. Cross-backbone tests and WritingPrompts provide bounded support across evaluation settings. Noise as a granularity axis lets a single level diffusion LM show planning like coarse to fine behavior without hierarchy or planner.

\section*{Limitations}
This work studies a focused mechanism question: whether diffusion noise can serve as a semantic granularity signal in a single-level diffusion LM. Our strongest evidence comes from the controlled topic-skeleton setting, while WritingPrompts is used only as a naturalistic health and segment-alignment check. One caveat is that early-recovery metrics parse partial outputs with macro-segment topic extraction, so span-contiguous outputs may be easier to score than scattered tokenwise outputs. BM-TOK controls reveal budget but not contiguity; early-order results partly reduce this concern. Also, reported values are two-seed means, so small cross-backbone differences, especially for SEDD, should be interpreted cautiously. Future work can test longer documents, learned semantic boundaries, richer groups, and combinations with explicit planners.

\bibliography{custom}

\appendix

\section{Method Details}
\label{app:ndgc_details}

\subsection{Group construction}
At noise level \(t\), \(B(t)\) is a target group length. The answer region is
partitioned in token order into
\begin{equation}
K(t)=\max(1,\lfloor L/B(t)\rfloor)
\end{equation}
contiguous groups. The first \(K(t)\ndminus 1\) groups use the target length
\(B(t)\), and the final group absorbs the remaining answer positions. Thus
\(B(t)\) is not a strict upper bound. We do not use a random offset or a
semantic boundary detector. The conditioning context \(c\) is not partitioned
or masked.

\subsection{Marginal preservation for structured exposure}
Structured exposure changes the correlation pattern of masking, not the
expected amount of masking. For each group, we first apply group masking with
probability
\begin{equation}
p_{\mathrm{span}}(t)=r(t)s(t).
\label{eq:p_span}
\end{equation}
If a group is not masked at the group level, each token in that group is
masked with residual probability
\begin{equation}
p_{\mathrm{token}}(t)
=
\begin{cases}
\dfrac{r(t)\ndminus p_{\mathrm{span}}(t)}
      {1\ndminus p_{\mathrm{span}}(t)}, & p_{\mathrm{span}}(t)<1,\\
0, & p_{\mathrm{span}}(t)=1.
\end{cases}
\label{eq:p_token}
\end{equation}
Then the expected masking probability of any answer token is
\begin{equation}
p_{\mathrm{span}}(t)
+
\bigl(1\ndminus p_{\mathrm{span}}(t)\bigr)p_{\mathrm{token}}(t)
=
r(t).
\label{eq:marginal_preservation}
\end{equation}
At high noise, \(s(t)\approx 1\), so masking is span correlated. At low noise,
\(s(t)\approx 0\), so \(p_{\mathrm{span}}(t)\approx 0\) and the process
reduces to tokenwise masking. At intermediate noise, group level correlation
and token level variation coexist.

\subsection{Reveal budget}
At each inference step, the policy selects up to \(b(t)\) answer tokens as
eligible positions:
\begin{equation}
b(t)
=
\max\!\left(
B(t),
\left\lfloor \frac{L}{T}\bigl(1.5\ndminus s(t)\bigr)\right\rfloor
\right).
\label{eq:reveal_budget}
\end{equation}
At high noise, this budget is close to \(B(t)\), so the policy attempts to
commit roughly one high confidence group. At low noise, the budget grows and
allows faster token level refinement. This schedule keeps denoising progress
comparable across systems. The budget matched tokenwise control uses the same
budget but commits positions independently, so it separates group commitment
from reveal pacing.

\subsection{Mixed commitment}
After group selection, tokens in selected groups are sampled from the model
induced reverse distribution. Since this distribution can assign probability
to \(\texttt{[MASK]}\), a selected token may remain masked after sampling.
Therefore \(b(t)\) controls eligible positions, not guaranteed revealed
positions. Tokens outside selected groups remain masked, and previously
revealed answer tokens are fixed.

\section{Dataset Construction and Filtering}
\label{app:data}

\subsection{\textsc{Synthetic-V4}}
\label{app:data_synthetic}

\textsc{Synthetic-V4} is a controlled benchmark for topic skeleton recovery. Each example is generated from an ordered latent topic skeleton. We first sample a topic set from a fixed topic bank. We then sample an order for the selected topics. This order defines the gold skeleton \(G=(g_1,\ldots,g_N)\). A prompt view is sampled next. We use three prompt views. The beats view gives short ordered story beats. The topic name view lists the topic names but does not give span boundaries. The weak view only asks for a coherent multi scene story. The target answer realizes the skeleton as \(N\) contiguous discourse spans. Each span is controlled by one topic. Optional transition templates connect neighboring spans. The model observes only the prompt. The gold skeleton, span boundaries, topic lexicon, and transition labels are saved as metadata and are used only by the evaluator.

\noindent\textbf{Prompt views.}
The beats view gives the strongest prompt signal because it shows ordered story beats. The topic name view gives the topic names but does not show span boundaries or transition labels. The weak view gives no topic names and asks only for a coherent multi scene story. This lets us evaluate both explicit and weak prompt settings while keeping the same hidden skeleton format.

\noindent\textbf{Target realization.}
The target answer is written as a sequence of contiguous spans. The number of spans equals the number of topics in the gold skeleton. Each span is tied to one gold topic and may contain transition text that links it to the next span. The target is therefore not a bag of topic words. A correct output must recover the right topic content, place topics in the right order, and keep local span content readable.

\noindent\textbf{Quality filtering.}
After generation, an example is kept only if it passes all checks. The span metadata must cover the answer region and match the sampled skeleton. Each topic span must contain enough lexical evidence for its assigned topic and must not be dominated by another topic. The answer must not be truncated. Repeated sentence patterns, special token leaks, replacement characters, and visible surface artifacts are removed. These filters are applied before train and test splits are formed.

\subsection{WritingPrompts}
\label{app:data_wp}

WritingPrompts is used as a natural long form check. Each example has a prompt and a reference story, but it does not provide a gold latent topic skeleton. We therefore do not use topic skeleton metrics on WritingPrompts. We use it only for output health and reference aligned segment similarity. The input prompt is kept visible and the answer region is denoised. The evaluation set, answer length limit, sampling steps, and nucleus setting are fixed across systems.

\section{Evaluation Metric Details}
\label{app:metric-details}

This appendix describes the evaluation metrics used in the controlled
Synthetic-V4 benchmark and the naturalistic WritingPrompts benchmark. The
main paper reports compact aggregate scores; here we specify how those
scores are computed. All thresholds are fixed before evaluation.

\subsection{Synthetic-V4 Topic-Sequence Extraction}
\label{app:synthetic-topic-extraction}

Each Synthetic-V4 example has a gold ordered topic skeleton
\[
G=(g_1,\ldots,g_N),
\]
where each topic corresponds to one contiguous discourse span in the
target passage. The gold topic sequence, topic lexicon, and span metadata
are used only by the evaluator and are not provided to the model at
generation time.

For each generated answer \(x\), the evaluator first splits the answer
into \(N\) equal-word macro-segments, where \(N\) is the number of gold
topics. Each macro-segment is then assigned a predicted topic by counting
topic-lexicon hits. Specifically, for segment \(j\) and topic \(g_k\), we
count how many words in the segment match the lexical set for \(g_k\).
The segment is assigned to the topic with the largest hit count. If a
segment contains no topic hit, it is assigned \textsc{none}.

This produces a raw predicted topic sequence
\[
\hat{Z}=(\hat{z}_1,\ldots,\hat{z}_N).
\]
For sequence-level metrics, \textsc{none} entries are removed:
\[
P=\mathrm{removeNone}(\hat{Z}).
\]
Duplicate topic predictions are retained rather than collapsed. Thus, if
a model repeats the same topic in multiple macro-segments, the repeated
topics remain in \(P\) and can hurt order and exact-match scores.

\subsection{Synthetic-V4 Final Skeleton Metrics}
\label{app:synthetic-final-metrics}

Let \(G\) be the gold topic skeleton and \(P\) be the compact predicted
topic sequence extracted from the generated answer.

\noindent\textbf{Final topic coverage.}
Final coverage measures whether the generated answer recovers the gold
topics, regardless of order:
\[
\mathrm{Coverage}(P,G)
=
\frac{
|\mathrm{set}(P)\cap \mathrm{set}(G)|
}{
|\mathrm{set}(G)|
}.
\]
This is reported as \textbf{Final coverage} or \textbf{Final cov.} in the
main tables.

\noindent\textbf{Order accuracy.}
Order accuracy measures whether the recovered topics appear in the gold
order. We compute the normalized longest-common-subsequence score:
\[
\mathrm{Order}(P,G)
=
\frac{\operatorname{LCS}(P,G)}{|G|}.
\]
This score is reported as \textbf{Order acc.}. It can be interpreted as
the fraction of the gold ordered skeleton recovered by the predicted
topic sequence.

\noindent\textbf{Exact skeleton match.}
Exact skeleton match is the strictest final recovery metric:
\[
\mathrm{Exact}(P,G)=\mathbf{1}[P=G].
\]
It is one only when the full compact predicted topic sequence exactly
matches the gold ordered skeleton, and zero otherwise.

\subsection{Synthetic-V4 Early Recovery}
\label{app:synthetic-early-recovery}

Final skeleton recovery does not by itself indicate whether a model forms
a plan early or repairs the output only near the end of denoising. We
therefore apply the same topic-sequence extractor to partially denoised
snapshots.

For the Synthetic-V4 main table, we evaluate reveal fractions
\[
\mathcal{R}=\{0.1,0.3,0.5,0.7\}.
\]
For each reveal fraction \(\rho\), let \(P_\rho\) be the predicted topic
sequence extracted from the partially denoised answer at that point. The
reported \textbf{Early rec. avg} score is
\[
\mathrm{EarlyRecAvg}
=
\frac{1}{|\mathcal{R}|}
\sum_{\rho\in\mathcal{R}}
\mathrm{Coverage}(P_\rho,G).
\]
The
reported \textbf{Early order. avg} score is
\[
EarlyOrderAvg =
\frac{1}{|R|}
\sum_{\rho \in \mathcal{R}}
Order(P_\rho, G).
\]
Thus, a higher Early rec. avg means that recognizable gold topics appear
earlier in the denoising trajectory. The trajectory figures additionally
plot ordered recovery over reveal fractions to show whether the early
topics appear in the correct order.
\begin{table*}[t]
\centering
\small
\caption{Full bad-output diagnostic decomposition on BD3LM \textsc{Synthetic-V4} (two-seed aggregate).
The high uniform-baseline bad-output rate is entirely explained by
repetition/diversity failures, not by non-repetition validity failures
such as empty outputs, token leakage, or control-character artifacts.}
\label{tab:bad-output-full-decomp}
\resizebox{\textwidth}{!}{%
\begin{tabular}{llcccccc}
\toprule
Category & Diagnostic rate $\downarrow$
& Baseline
& Struct.\ commit.\ only
& Struct.\ exposure only
& Tokenwise early
& Budget-matched tok.
& NDGC \\
\midrule
Validity
& Empty / too short
& 0.000 & 0.000 & 0.000 & 0.000 & 0.000 & 0.000 \\
Validity
& Token leakage
& 0.000 & 0.000 & 0.000 & 0.000 & 0.000 & 0.000 \\
Validity
& Replacement-char.\ artifact
& 0.000 & 0.000 & 0.000 & 0.000 & 0.000 & 0.000 \\
Validity
& Extreme length
& 0.000 & 0.012 & 0.000 & 0.000 & 0.000 & 0.000 \\
Validity
& Control-char.\ artifact
& 0.000 & 0.000 & 0.000 & 0.000 & 0.000 & 0.000 \\
\midrule
Repetition/diversity
& Low unique-token ratio
& 0.997 & 0.006 & 0.000 & 0.000 & 0.000 & 0.000 \\
Repetition/diversity
& High repeated 4-gram rate
& 0.000 & 0.000 & 0.000 & 0.000 & 0.000 & 0.000 \\
Repetition/diversity
& High repeated 2--5gram count
& 0.996 & 0.878 & 0.198 & 0.083 & 0.000 & 0.030 \\
Repetition/diversity
& Additional 5-gram/sent./line rep.
& 0.000 & 0.000 & 0.000 & 0.000 & 0.070 & 0.007 \\
\midrule
Grouped union
& Non-rep.\ validity failure
& 0.000 & 0.012 & 0.000 & 0.000 & 0.000 & 0.000 \\
Grouped union
& Rep./diversity failure
& 0.997 & 0.878 & 0.198 & 0.083 & 0.070 & 0.037 \\
Grouped union
& Bad-output union
& 0.997 & 0.890 & 0.198 & 0.083 & 0.070 & 0.037 \\
\bottomrule
\end{tabular}%
}
\vspace{2pt}
\begin{minipage}{0.98\textwidth}
\footnotesize
\textit{Note.}
The non-repetition validity failure group is the union of
empty/too-short output, token leakage, replacement-character artifacts, extreme-length output, and control-character artifacts.
The ``Additional 5-gram/sent./line rep.'' row counts additional-only repetition/diversity failures triggered by high repeated-5-gram rate or sentence/line-level repetition that are not already captured by the preceding displayed repetition/diversity rows.
Thus the displayed repetition/diversity block is a self-contained decomposition of the grouped repetition/diversity union.
The reported bad-output union is the union of all validity and repetition/diversity triggers.
\end{minipage}
\end{table*}
\subsection{Output-Health Metrics}
\label{app:output-health}

Bad-output rate is the fraction of generations that trigger at least one
severe validity or degeneration flag. This metric is not intended to
measure planning. It is included to ensure that skeleton or coherence
scores are not explained by degenerate generations.

For both benchmarks, an output is marked as bad if any of the following
conditions holds:
\begin{itemize}
\setlength{\itemsep}{0pt}
\setlength{\parsep}{0pt}
\setlength{\topsep}{2pt}
    \item it contains fewer than five word tokens;
    \item it visibly leaks mask, padding, unknown, separator, or other
    special tokens;
    \item it contains replacement-character artifacts;
    \item its control-character rate is greater than \(0.02\);
    \item its length ratio relative to the expected answer length is
    below \(0.05\) or above \(6.0\);
    \item its repeated 4-gram rate is greater than \(0.65\);
    \item any repeated 2--5gram appears at least 20 times;
    \item it has at least 20 word tokens but a unique-token ratio below
    \(0.15\).
\end{itemize}

For WritingPrompts, we also mark severe repetition when the repeated
5-gram rate is greater than \(0.60\), the repeated-sentence rate is
greater than \(0.50\), or the repeated-line rate is greater than \(0.50\).
The reported bad-output rate is the union of these predeclared flags.

\noindent\textbf{Repeated 4-gram rate.}
The repeated 4-gram rate reported for WritingPrompts is
\[
\mathrm{Rep}\text{-}4(x)=1-\mathrm{distinct}\text{-}4(x),
\]
where \(\mathrm{distinct}\text{-}4(x)\) is the number of unique word-level
4-grams divided by the total number of word-level 4-grams. Lower values
indicate less local repetition.

\subsection{WritingPrompts RASC-Best}
\label{app:rasc-best}

WritingPrompts does not provide a gold latent topic skeleton, so we do
not compute topic-skeleton recovery on this benchmark. Instead, we use
reference-aligned segment coherence as a naturalistic diagnostic of
long-form semantic alignment.

For each generated story \(x\) and reference story \(y\), we split both
texts into sentence-balanced macro-segments, targeting up to six
segments:
\[
x\rightarrow(s_1,\ldots,s_M),
\qquad
y\rightarrow(r_1,\ldots,r_L).
\]
The generated and reference stories are segmented independently. 
The table reports RASC-best:
\[
\mathrm{RASC}_{\mathrm{best}}(x,y)
=
\frac{1}{M}
\sum_{i=1}^{M}
\max_{1\leq j\leq L}
\cos(e(s_i),e(r_j)).
\]
Each segment is encoded with a fixed MiniLM sentence encoder
\citep{wang2020minilm}, and segment similarity is computed by cosine
similarity between normalized segment embeddings. RASC-best allows moderate differences in segment length and
ordering, so it should be interpreted as a naturalistic coherence and
semantic-alignment diagnostic rather than direct evidence of latent
skeleton recovery.

\subsection{Aggregation Across Seeds}
\label{app:seed-aggregation}

Unless otherwise stated, reported table values are means over two
fine-tuning seeds. Within each seed, all compared systems are evaluated
on the same set of examples. We first average each metric over examples
within a seed and then average the resulting seed-level means.

For pairwise comparisons, we compute deltas on matched examples within
each seed before averaging across seeds. 
\subsection{Bad-Output Trigger Thresholds and Diagnostic Rates}
\label{app:bad-output-triggers}

The main paper reports a compact decomposition of bad outputs into
non-repetition validity failures and repetition/diversity failures. This
appendix gives the full trigger thresholds and per-diagnostic trigger
rates. All thresholds are fixed before evaluation.

\noindent\textbf{Trigger thresholds.}
\label{app:trigger}
An output is marked as bad if it triggers at least one of the following
conditions:
\begin{itemize}
\setlength{\itemsep}{0pt}
\setlength{\parsep}{0pt}
\setlength{\topsep}{2pt}
    \item \textbf{Empty or too short:} fewer than five word tokens.
    \item \textbf{Token leakage:} visible mask, padding, unknown,
    separator, sentence-boundary, or other special tokens.
    \item \textbf{Replacement-character artifact:} at least one Unicode
    replacement character.
    \item \textbf{Extreme length:} generated-to-expected word-count ratio
    below \(0.05\) or above \(6.0\).
    \item \textbf{Control-character artifact:} control-character rate
    above \(0.02\), excluding newline and tab.
    \item \textbf{Low lexical diversity:} at least 20 word tokens and
    unique-token ratio below \(0.15\).
    \item \textbf{High repeated 4-gram rate:} repeated 4-gram rate above
    \(0.65\), where repeated 4-gram rate is \(1-\mathrm{distinct}\text{-}4\).
    \item \textbf{High repeated n-gram count:} some repeated 2--5gram
    appears at least 20 times.
\end{itemize}

\end{document}